\documentclass[runningheads]{llncs}
\usepackage{graphicx}
\usepackage{comment}
\usepackage{amsmath,amssymb} % define this before the line numbering.
\usepackage{color}
\usepackage{newunicodechar}

\begin{document}
% \linenumbers
\pagestyle{headings}
\mainmatter
\def\ECCVSubNumber{683}  % Insert your submission number here

\title{Improving Skeleton-based Action Recognition \\with Robust Spatial and Temporal Features} % Replace with your title

%\begin{comment}
% \titlerunning{ECCV-20 submission ID \ECCVSubNumber} 
% \authorrunning{ECCV-20 submission ID \ECCVSubNumber} 
% \author{Anonymous ECCV submission}
% \institute{Paper ID \ECCVSubNumber}
%\end{comment}
%******************

%\begin{comment}
\titlerunning{Abbreviated paper title}
\author{Zeshi Yang \and
Kangkang Yin}
\authorrunning{}
\institute{Simon Fraser University}
%\end{comment}

\definecolor{myOrange}{rgb}{1,0.3,0}
\definecolor{myGreen}{rgb}{0,0.8,0.5}
\definecolor{kkBlue}{rgb}{0,0.0,0.0}
% \definecolor{myOrange}{rgb}{0,0,0}
% \definecolor{myGreen}{rgb}{0,0.,0.}
% \definecolor{kkBlue}{rgb}{0,0.0,0.}

\newcommand{\zeshi}[1]{\textcolor{myOrange}{#1}}
\newcommand{\comm}[1]{\textcolor{myGreen}{#1}}
\newcommand{\kk}[1]{\textcolor{kkBlue}{#1}}

%******************
\maketitle
\begin{abstract}
% \kk{Recently skeleton-based action recognition has made significant progresses in the computer vision community. Most state-of-the-art algorithms are based on Graph Convolutional Networks (GCN), and target at improving the GCN backbone layers. In this paper, we propose to alter the last few layers of the network architecture instead, including the global pooling layer and the fully connected layers. We show that the recognition accuracy can be improved by a large margin when these last few layers can help extract temporal information from features computed by the GCN backbone. We also formally advocate the use of direction-invariant joint coordinates as input features to neural networks to further improve the recognition performance. Few prior work has used Direction-Invariant Features (DIF) in a principled way. We demonstrate the effectiveness of our ideas by comparing the results with those of ST-GCN and AGCN on three datasets: NTU-RGBD60, NTU-RGBD120 and  Kinetics-Skeleton.}
\kk{Recently skeleton-based action recognition has made significant progresses in the computer vision community. Most state-of-the-art algorithms are based on Graph Convolutional Networks (GCN), and target at improving the network structure of the backbone GCN layers. In this paper, we propose a novel mechanism to learn more robust discriminative features in space and time. More specifically, we add a Discriminative Feature Learning (DFL) branch to the last layers of the network to extract discriminative spatial and temporal features to help regularize the learning. We also formally advocate the use of Direction-Invariant Features (DIF) as input to the neural networks. We show that action recognition accuracy can be improved when these robust features are learned and used. We compare our results with those of ST-GCN and related methods on four datasets: NTU-RGBD60, NTU-RGBD120, SYSU 3DHOI and Skeleton-Kinetics.}

\end{abstract}

\section{Introduction}

\kk{Human action recognition is a challenging task. In computer vision, there are a large body of research investigating action recognition directly from video inputs. Skeleton-based action recognition, however, works with extracted position and/or orientation of skeletal joints to model the dynamics of human motion. Compared with RGB images, skeletal information is more robust to illumination changes and scene variations. Therefore, skeleton-based action recognition algorithms can potentially well complement video-based recognition methods.}

\kk{Skeleton-based action recognition has made great progresses recently with the adoption of Graph Convolutional Networks (GCN). For example, Spatial Temporal GCN (ST-GCN) constructs a set of spatial temporal graph convolutions on skeleton sequences, and achieves state-of-the-art results with a few variations ~\cite{yan_2018_AAAI,Shi_2019_CVPR,Shi_2019_CVPR2,si_2019_CVPR,li_2019_arxiv_sym}. However, almost all GCN-based methods focus on designing the backbone graph convolutional layers and share almost identical last layers. The last layers, which we will call the global branch from now on, consist of a Global Average Pooling (GAP) layer followed by a fully connected layer. The GAP layer aggregates information across both the spatial and temporal domain, and thus is effective at reducing overfitting and improving robustness wrt temporal translations. However, it also removes rich information in the full feature maps that we may utilize for better classification and generalization.}% .............. is good at capturing general motion features. However, it may restrict the neural network's generalization ability. Global average pooling layer aggregates information across the whole spatial and temporal domain, which make it may use specific data distribution pattern in dataset to fit training examples. 

\begin{figure}[tbh]
\centering
\includegraphics[scale = 0.05]{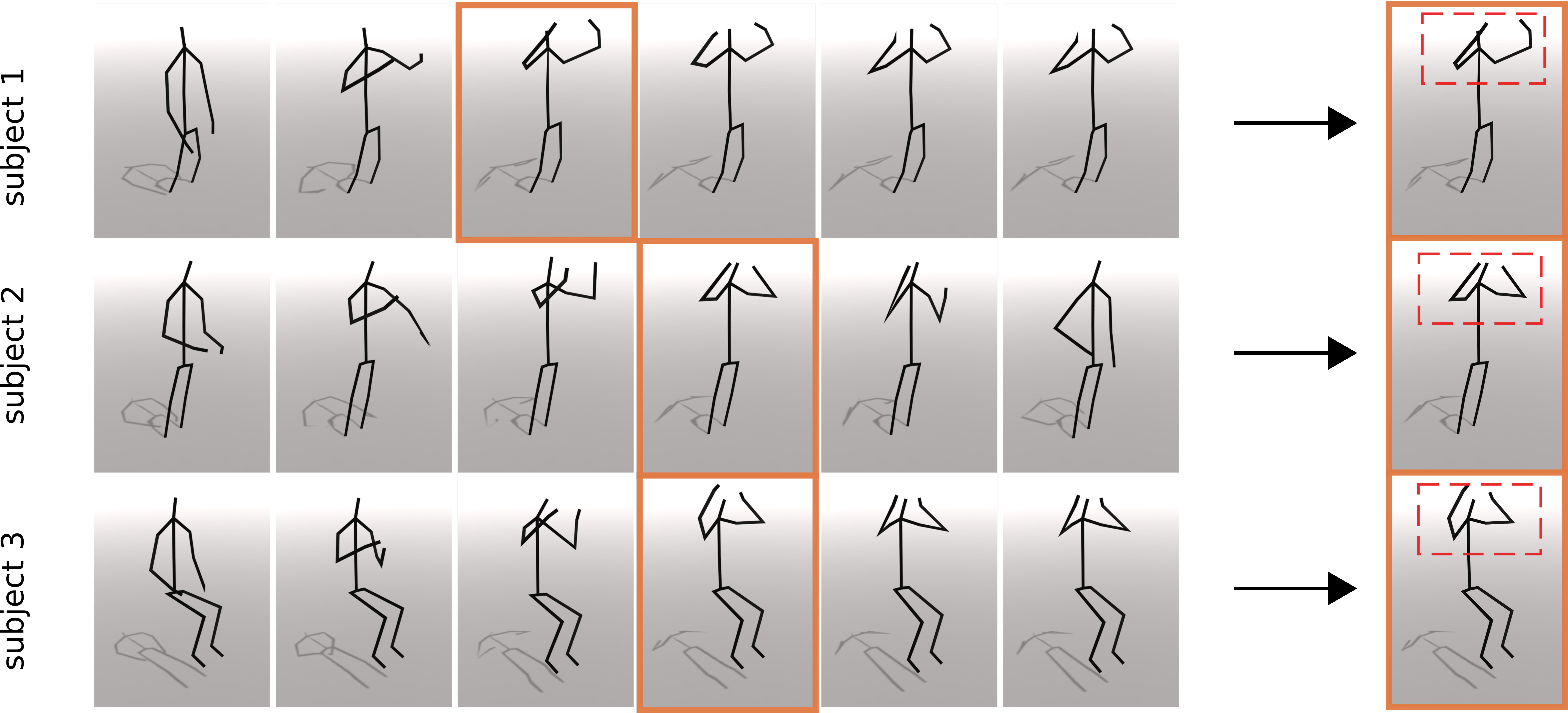}
\caption{"put on headphone" action performed by three different subjects in NTU-RGBD dataset. }
\label{fig:headphone}
\end{figure}

\kk{We augment the global branch with a Discriminative Feature Learning (DFL) branch to better utilize the rich information contained in the full feature maps extracted by the GCN backbone. Human skeleton sequences involve complex dynamics. Same actions performed by different subjects or even the same subjects may vary in style and speed. In order to obtain classification models that are robust to these non-essential motion variations, we need to extract features that are insensitive to factors such as different action styles or different camera settings. In another word, we need to extract features that can capture the essence of an action. For instance, the three ``put on headphone'' action clips in Figure~\ref{fig:headphone} all have similar motion semantics: in the temporal domain, they all have one motion segment where the hands raise up and approach the head; in the spatial domain, the arm joints move in similar ways while the lower body joints can be quite different. If we could extract these distinctive features in space and time for a particular type of action, we would be able to recognize that type of action with more success.}

\kk{Our DFL branch help the GCN backbone extract \textit{essential} and \textit{robust} discriminative features by first segmenting the full feature maps in time. Motion primitives can thus be examined independently in time rather than averaged together. Interestingly, distinctive spatial characteristics can be learned too, once features are separated in time. We train the GCN backbone together with both the global branch and the DFL branch end to end, and fuse classification results of both branches together at the inference stage. Our model is simple to implement and generates comparable or better results on testing datasets.}

\kk{Another non-essential motion variation is where the character faces. That is, a walk to the south and a similar walk to the north should be classified as the same type of motion. It is well-known in the computer animation community that direction invariance is desirable~\cite{Ma_2019_PG}. However, most action recognition systems in the computer vision community use absolute joint coordinates as input, so motions towards different directions look quite different to the neural networks. Therefore it is hard to learn direction-invariant classifiers with absolute input features, unless the dataset is big enough to include example motions of all directions. We therefore advocate transforming input skeletal features into Direction Invariant Features (DIF) for easy training and learning of more robust classifiers. DIF features can be extracted easily in a data preprocessing stage.}

\kk{The major contributions of this work include: 1) a simple and novel neural network structure which facilitates the GCN backbone to learn more robust discriminative features. 2) adopting DIF features to obtain more robust direction-invariant classifiers. 3) ablations and comparisons done on four datasets, where we achieve comparable or better performance.}
\section{Related Work}
\subsubsection{Skeleton-based Action Recognition}
\kk{Traditional skeleton-based action recognition methods heavily rely on handcrafted features~\cite{Fernando_2015_CVPR,vemulapall_2014_CVPR}. In contrast, deep learning methods extract features automatically and can achieve better results. There are mainly three categories of deep learning methods for skeleton-based action recognition: CNN-based methods, RNN-based methods, and GCN-based methods. CNN-based methods apply convolutions on pseudo-images formed by skeleton sequences~\cite{kim_2017_CVPRW,liu_2017_arxiv,ke_2017_CVPR,liu_2018_CVPR,li_2017_ICMEW,li_2017_ICMEW_2,li_2018_arxiv_co}. RNN-based methods treats skeleton data as vector sequences~\cite{Shahroudy_2016_CVPR,liu_2016_ECCV,song_2017_AAAI,zhang_2017_CVPR,Cao_2017_CVPR,li_2018_arxiv_skeleton}. Neither CNN-based nor RNN-based methods take the graph nature of human pose into consideration.}

\kk{Most state-of-the-art skeleton-based recognition methods are GCN-based. ST-GCN directly models the skeleton data as a pre-defined spatial temporal graph~\cite{yan_2018_AAAI}. The graph can also be learned adaptively~\cite{Shi_2019_CVPR2,Li_2019_CVPR,peng_2019_AAAI}. Most GCN-based methods focus on improving the GCN backbone layers, while we only alter the last few layers and keep the GCN backbone unchanged. GCN can also be used as building blocks for LSTM models~\cite{si_2019_CVPR}. It is also common to integrate skeleton-based methods with video-based methods to obtain two-stream models that can further improve the classification accuracy~\cite{shi_2019_arxiv,liu_2019_TPAMI}. In this paper, we only focus on the single stream of skeleton input.}
% \kk{\cite{wang2016temporal}: a)	Also segment sequences in time domain, while in data preprocessing part; b)	Only use segmentation features to classify actions, no global average pooling features mare used; c) We want to avoid over-fitting, while TSN is designed for classify long term motions}

% \kk{\cite{zhang2017alignedreid}: a) Also use segmented features and global average pooling feature at the same time, while in spatial domain; b) Use DP to compute align distance; c) Have the same phenomenon: with the help of segmented feature, global average pooling feature can be improved: explanations are marked with yellow color}

% \kk{\cite{sun2018beyond}: a)	also use equally segmented features to classify actions, while without global average pooling feature and in spatial domain; b) use adaptive partition strategy to refine results(what we want to do after siggraph}

\subsubsection{\kk{GAP and Temporal Segmentation}}

% \kk{\cite{tang_2018_ICCV} also investigates the temporal domain to improve performance.}

% \kk{\cite{martinez_2019_ICCV}: a)	First paper to argue about weakness of global pooling; b)	Focus on fine grained action classification(in spatial domain)}
\kk{GAP layer has its pros and cons. Some previous work, which directly inspired our work, tried to deal with GAP's limited ability to model local spatial and temporal characteristics. \cite{martinez_2019_ICCV} proposed to use a bank of filters to differentiate fine-grained differences in actions. \cite{wang_2016_ECCV} proposed a Temporal Segment Network(TSN) that segments input video clips into multiple parts. \cite{tang_2018_ICCV} used deep reinforcement learning to distil informative frames from input clips to GCN. Both \cite{wang_2016_ECCV} and \cite{tang_2018_ICCV} segment raw input, while our method segments features computed by the backbone GCN. In person re-identification, some work partitions features in the spatial domain to select specific parts of the human body to boost performance~\cite{sun_2018_ECCV,zhang_2017_arxiv}. We only partition the extracted feature in the temporal domain, and a spatial attention mechanism can be learned implicitly by the GCN backbone layers.}

\subsubsection{\kk{Direction Invariant Features}}
% \kk{There are a few papers that do use direction invariant  features~\cite{Shi_2019_CVPR2,Shi_2019_CVPR}. Their features are not exactly the same as ours, but similar. Furthermore, their direction invariance is not discussed and the feature calculation is not well formulated.}
\kk{Within the skeleton-based action recognition literature, some works use raw joint positions as input~\cite{yan_2018_AAAI}; some transform joint positions to a semi-local coordinate system which is close to character~\cite{Shi_2019_CVPR,Shi_2019_CVPR2,Li_2019_CVPR}. We adopt the principled way of transforming input features into  Direction Invariant Features(DIF) described in~\cite{Ma_2019_PG}. Our experiments show that this simple preprocessing alone can boost performance by a large margin.}

\section{Methods}
\label{sec:method}

\subsection{Spatial Temporal Graph Convolutional Networks}
\kk{We briefly summarize the spatial-temporal graph convolutional networks (ST-GCN) developed by Yan et al.~\cite{yan_2018_AAAI} for skeleton based action recognition. Due to space limit, we refer readers to~\cite{yan_2018_AAAI} for more details not mentioned here.}

\begin{figure}[tbh]
  \begin{minipage}[c]{0.55\textwidth}
  \centering
    \includegraphics[scale = 0.06]{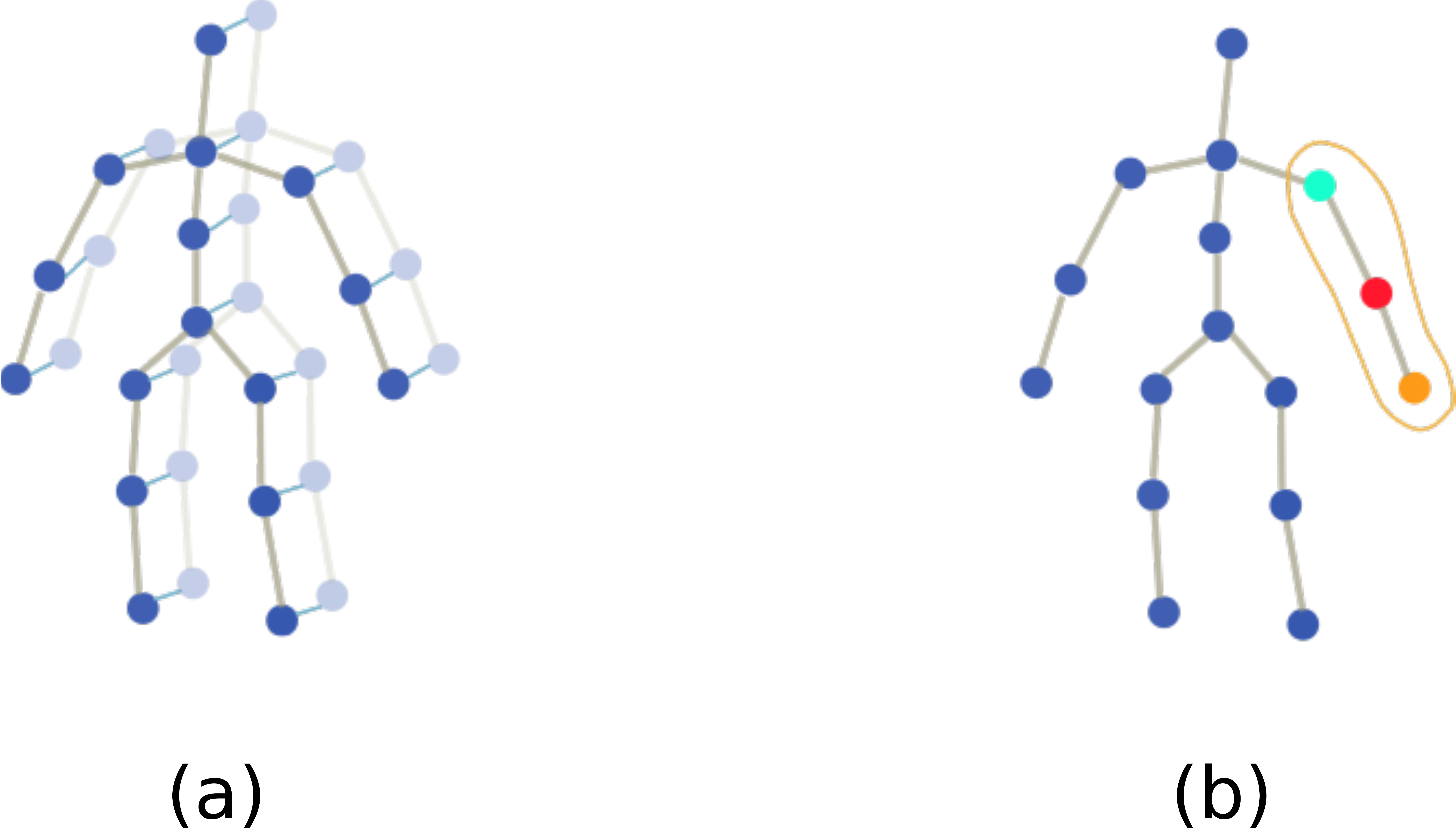}
  \end{minipage}\hfill
  \begin{minipage}[c]{0.40\textwidth}
  \centering
    \caption{
       Spatial-temporal graph (a) and the receptive field of graph convolutions (b).
    } \label{skel}
  \end{minipage}
\end{figure}

\kk{Figure~\ref{skel}(a) shows the spatial temporal graph constructed from 3D joint positions: vertices represent joints; edges in the same frame connect adjacent joints; and edges across frames connect the same joint in consecutive frames. We use the so-called spatial configuration partitioning strategy for convolution operations as shown in Figure \ref{skel}(b). For a vertex $v_{i}$ colored red in the figure, its centripetal neighbor colored green and centrifugal neighbor colored orange form the receptive field.}

%\begin{figure}[tbh]
%\centering
%\includegraphics[scale = 0.06]{images/graph.pdf}
%\caption{Spatial-temporal graph (a) and the receptive %field of graph convolutions (b).}
%\label{skel}
%\end{figure}

\kk{The convolution for a single frame in the spatial domain is formulated as:
\begin{equation}
    \mathbf{f}_{out}(v_{i}) = \sum_{v_{j} \in B(v_{i})} \frac{1}{Z_{i}(v_{j})}\mathbf{f}_{in}(v_{j}) \ast \mathbf{w}(l_{i}(v_{j}))
\end{equation}
where $\mathbf{f}$ is the feature map. $B(v_{i})$ is the 1-distance receptive field where convolution is applied to aggregate features. $Z$ is the cardinality of $B$. $\mathbf{w}$ is the weight function. $l$ is the partition function. The convolution in the temporal domain is simply a 1D convolution of the same joint along the temporal axis.
}
\subsection{Our Pipeline}
\begin{figure}[tbh]
\centering
\includegraphics[scale = 0.040]{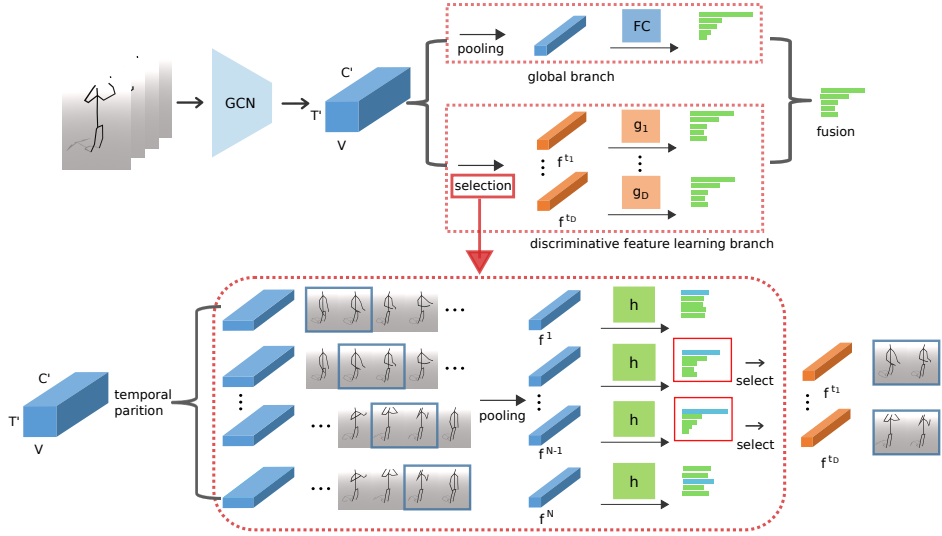}
\caption{Architecture of our model. Features extracted by the GCN backbone go through two classification branches. The global branch aggregates features by averaging them in space and time. The discriminative feature learning branch examines features in shorter temporal segments and selects discriminative features via training for classification.}
\label{pipeline}
\end{figure}

\kk{Figure~\ref{pipeline} shows the architecture of our model. We use the same GCN  backbone layers as~\cite{yan_2018_AAAI}. After the features, represented by the blue block in the figure, are extracted by the backbone layers, they are fed to two classification branches: a \textit{global} branch and a \textit{discriminative feature learning} (DFL) branch. The global branch is exactly the same as the GAP layer and the successive fully connected layer in~\cite{yan_2018_AAAI,Shi_2019_CVPR2}. The DFL branch helps extract robust features in both space and time through training. These features focus on the essence of actions, rather than spatial and temporal variations caused by different subjects, styles, and camera settings. The two branches are jointly trained end to end, and their classification outputs are combined together as the final result.} 

\subsection{Global Branch}
\kk{The global branch follows the common skeleton-based action recognition pipeline such as described in~\cite{yan_2018_AAAI,Shi_2019_CVPR2}. Denote the input features to the GCN backbone as $\textbf{f}_{in}$, whose size is $C \times T \times V$. $C$ is the dimension of the joint coordinates, i.e., three dimensions using 3D joint positions. $T$ is the temporal dimension, i.e., the number of frames. $V$ is the spatial dimension, i.e., the number of joints. Denote the feature map computed by the GCN backbone as $\textbf{f}_{out}$, whose size is $C' \times T' \times V$. A pooling layer then averages $\textbf{f}_{out}$ in space and time to obtain an aggregated feature $\textbf{f}$ of dimension $C'$. Then a fully connected layer linearly classifies $\textbf{f}$ into scores $\mathbf{\hat{y}}^g \in \mathbb{R}^{c}$, where $c$ is the number of action classes. We use the standard softmax cross entropy loss in training:
\begin{equation}
    \mathcal{L}_{g} =  \sum_{k = 1}^{c}-y_{k}\ log \frac{e^{\hat{y}_{k}^{g}}}{\sum_{l=1}^{c}e^{\hat{y}_{l}^{g}}}
\end{equation}
where $\{y_{k}\}$ denotes the one-hot vector corresponding to the ground truth labels, and $\mathbf{\hat{y}}^g=\{\hat{y}_{k}^{g}\}$ the network output.}

\subsection{Discriminative Feature Learning Branch}
\kk{The DFL branch first segments the input in the temporal domain. That is, the extracted feature map $\mathbf{f}_{out}$ is uniformly segmented into $N$ fragments in time and then averaged by a pooling layer into features $\mathbf{f}^{i}$. We then train a shared classifier $\mathbf{h}$ using a fully connected layer which outputs predicted class scores $\mathbf{\hat{y}}^{i} \in \mathbb{R}^{c}$, $i = 1,... N$. The total loss of all segments is:}
\begin{equation}
    \mathcal{L}_{s} =\frac{1}{N} \sum_{i = 1}^{N}\sum_{k = 1}^{c}-y_{k}\ log\frac{e^{\hat{y}_{k}^{i}}}{\sum_{l=1}^{c}e^{\hat{y}_{l}^{i}}}
\end{equation}
We also define a saliency score for each $\mathbf{f}^{i}$ as the maximum score among $c$ classes:
\begin{equation}
    score(\mathbf{f}^{i}) = max _{j \in  \left\{1,...,c \right\}}\mathbf{\hat{y}}_{j}^{i}
\end{equation}
\kk{Then we select $D$ features with the top saliency scores as the most discriminative segments for the associated full feature map. We sort these features in time as $\mathbf{f}_{d} = \left\{\mathbf{f}^{t_1},... \mathbf{f}^{t_D} \right\}$. According to our experiments and ablation studies, $\mathbf{f}_{d}$'s of actions in the same category are roughly aligned and correspond well to the semantic meaning of the actions. For example,  almost all ``put on headphone'' actions start from a segment of ``raising hands'' followed by a segment of ``putting headphone on the head''.}

\kk{After the discriminative features are selected, we assign a classifier $\mathbf{g}_{m}$ for each discrimnative feature $\mathbf{f}^{t_m}$. Note that unlike $\mathbf{h}$, the parameters of $\mathbf{g}_{m}$ are not shared, which means each classifier can focus on features of shorter action primitives. For example, classifier $\mathbf{g}_{m}$ only uses feature $\mathbf{f}_{j}^{t_m}, j = 1...,n$, where $n$ is the number of training clips in a batch. We denote the output of the classifier $\mathbf{g}_{m}$ as $\mathbf{\hat{y}^{m}} =\{\hat{y}_{k}^{m}\}= \mathbf{g}_{m}(\mathbf{f}^{t_m})$. The loss function for all the $\mathbf{g}_{m},m=1,...,D$ is:
\begin{equation}
    \mathcal{L}_{d} = \frac{1}{D}\sum_{m = 1}^{D}\sum_{k = 1}^{c}-y_{k}\ log \frac{e^{\hat{y}_{k}^{m}}}{\sum_{l=1}^{c}e^{\hat{y}_{l}^{m}}}
\end{equation}
We then sum up the classification results $\mathbf{\hat{y}}^{m}$ together to form an aggregated classification result $\mathbf{\hat{y}}^{a} =\{\hat{y}_{k}^{a}\}= \sum_{m=1}^{D}\mathbf{\hat{y}}^{m}$. The corresponding classification loss is defined as:
\begin{equation}
    \mathcal{L}_{a} =  \sum_{k = 1}^{c}- y_{k}\ log \frac{e^{\hat{y}_{k}^{a}}}{\sum_{l=1}^{c}e^{\hat{y}_{l}^{a}}}
\end{equation}
}

\subsection{Fusion of Two Branches}
\kk{The global branch and the DFL branch are jointly trained end to end. The total loss function for the two branches trained together is:
\begin{equation}
    \mathcal{L} = \mathcal{L}_{g} + \mathcal{L}_{s} + \mathcal{L}_{d} + \mathcal{L}_{a}
\end{equation}
In the inference stage, we fuse outputs from both branches to compute the final classification result: 
\begin{equation}
    \mathbf{\hat{y}} = 0.5*(softmax(\mathbf{\hat{y}}^{g})+ softmax(\mathbf{\hat{y}}^{a}))
\end{equation}
}

\begin{figure}[tbh]
\centering
\includegraphics[scale = 0.028]{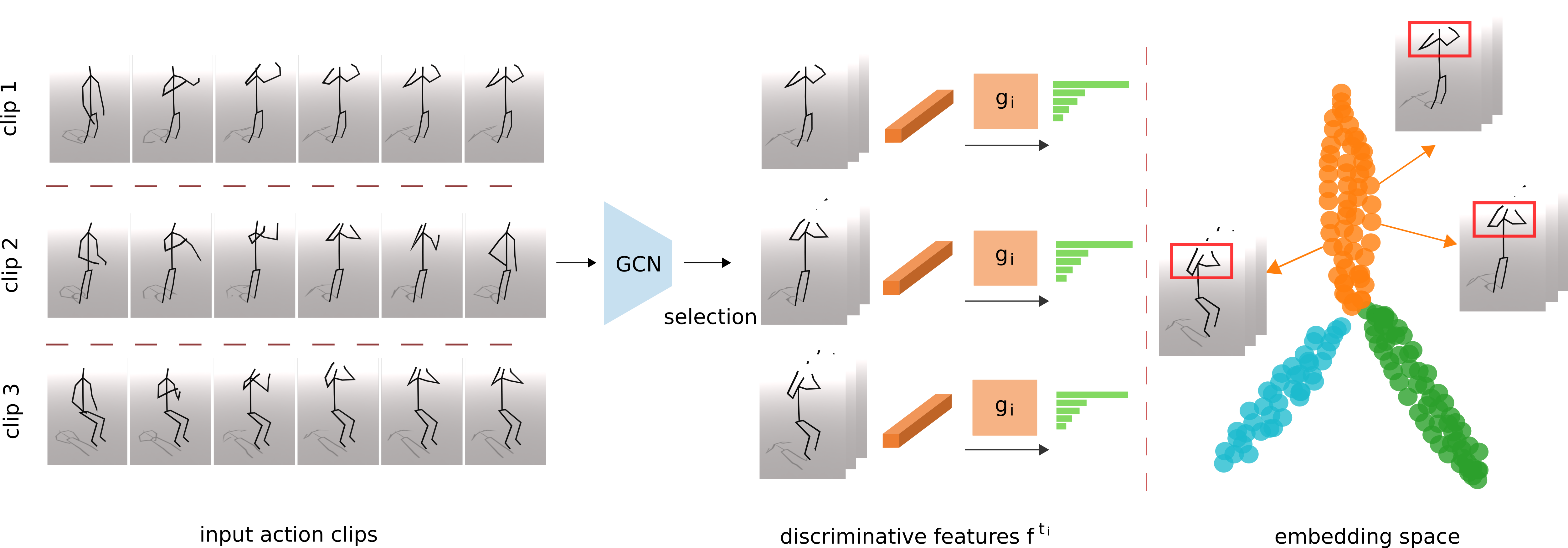}
\caption{Conceptual illustration of the discriminative feature learning branch. Three sample clips from the ``put on headphone'' action illustrate the input. For each sample a group of discriminative features are selected and sorted in the temporal domain. Here we only visualize feature $\mathbf{f}^{t_i}$ with temporal index $t_i$, which are then sent to classifier $\mathbf{g}_i$. The selected features are clustered during training into a low-dimensional embedding space, which is 2D here for ease of conceptual visualization. Points of different color represent different action types. The learned features tend to focus more on semantically important joints, which are the upper body joints for the ``put on headphone'' action.} % to have a compact representation in embedding space. }
\label{robust}
\end{figure}

\kk{With the DFL branch, the GCN backbone will learn more discriminative features both in space and time. In the temporal domain, classifier $\mathbf{g}_{m}$ can only see selected features $\mathbf{f}^{t_m}$, which makes its classification less sensitive to changes in other segments of the feature maps. Thus the temporal segmentation can help to learn temporally robust features. Another consequence of temporal segmentation is that only feature segments with similar semantics are examined together during training. The backbone network will therefore be driven to focus more on semantically important joints for these segments and ignore spatial variations that are not essential for the actions. Thus the temporal segmentation can also help to learn spatially robust features. Figure~\ref{robust} illustrates how the temporal segmentation can induce learning of both spatially and temporally robust features. We will validate this mechanism further in Section~\ref{sec:visualization} where we visualize the learned discriminative feature maps.}

\section{Experiments}
\subsection{Datasets}
\label{sec:datasets}
\kk{We train our model with the training set from the following datasets, and report the top-1 accuracy on the validation set of the corresponding datasets.}

\kk{\textbf{NTU-RGBD 60} is a commonly used indoor action recognition dataset~\cite{Shahroudy_2016_CVPR}. It contains 56880 action clips in 60 classes performed by 40 different subjects. The dataset provides 3D positions of human joints captured by Kinect depth sensors. Each action is captured by 3 cameras with horizontal angles at $-45^{\circ}$, $0^{\circ}$ and $45^{\circ}$. Each subject has 25 joints, and there are at most two subjects in any sequence. We follow two testing benchmarks: Cross-Subject (X-Sub) and Cross-View (X-View). In the cross-subject setting, 40320 clips are used for training and 16560 clips for testing, and subjects are different in the two subsets. In the cross-view setting, videos captured by camera 2 and 3 are used for training (37920 clips) and the ones captured by camera 1 for validation (18960 clips).}

\kk{\textbf{NTU-RGBD 120} is currently the largest captured indoor action dataset~\cite{liu_2019_TPAMI}, which builds upon the above mentioned NTU-RGBD 60 dataset. It contains $114480$ videos in 120 classes performed by 106 subjects of diverse demographic characteristics. The recording angles remain the same as those of NTU-RGBD 60, while more camera setups, such as different heights and distances, are included. The dataset also provides two benchmarks: Cross-Subject (X-Sub), and Cross-Setup (X-Setup). In the cross-subject benchmark, videos performed by half of the subjects are used for training, and the others for testing. In the cross-setup benchmark, videos captured with half of the camera settings are used for training, and the others for validation. In addition, we also use the cross-view benchmark proposed in NTU-RGBD 60. NTU-RGBD 120 is by far the most challenging captured skeleton dataset in terms of size and variety.}

\kk{\textbf{SYSU} dataset contains 12 types of human-object interactions such as activities using chairs or cups. There are 480 video clips in total performed by 40 subjects. We use the second benchmark settings which uses half of the subjects for training and the others for testing.}

\kk{\textbf{Kinetics-Skeleton} is derived from the Kinetics dataset~\cite{kay_2017_kinetics}, which contains approximately 300000 videos of 400 classes sourced from Youtube. Kinetics-Skeleton contains 2D joint positions and their corresponding confidence scores extracted from the Kinectics videos using the OpenPose toolbox~\cite{Cao_2017_CVPR}. 240000 clips are used for training and 20000 clips for validation. Since the extracted joint positions are extremely noisy and the top-1 accuracy is very low, we also report the top-5 accuracy on this dataset.}

\subsection{DIF Feature Calculation and Data Preprocessing}
\kk{Various input features have been employed in the literature, such as global joint positions~\cite{yan_2018_AAAI}, and joint positions transformed into a semi-local coordinate system where the origin is defined by the character's position in the first frame of the clip, and the axes defined by the shoulder joints and the gravitational direction~\cite{Shi_2019_CVPR2}. We follow the principled way of calculating Direction Invariant Features (DIF) proposed in ~\cite{Ma_2019_PG}, which outperforms global features for various tasks in character animation.} 

\kk{More specifically, we first define a local coordinate system whose origin is located at the spine joint of the character. The $Y$ axis is the unit vector pointing from the spine joint to the  chest joint. The $X$ axis is the cross product of $Y$ and the vector pointing from the left shoulder joint to the right shoulder joint. Finally the $Z$ axis is the cross product of $X$ and $Y$. We then convert all joint positions from the camera coordinate system into the above properly defined local coordinate system. The transformed features are invariant to the orientation of the character's action, and help increase the generality of models trained on small datasets. We could instead design a data augmentation method that replicates actions to different directions to help the learning of direction robust models, but such an approach would unavoidably be complex and costly.}

\kk{We transform the original features into DIF features in a preprocessing stage. In addition, we normalize motion clips into uniform length (100 frames) by linear interpolation, when a dataset contains clips of various length. For the Kinetics-Skeleton dataset, we also follow the data augmentation procedure described in~\cite{yan_2018_AAAI}, which involves randomly translating and rotating characters in randomly selected fragments from the original clips.}

\subsection{Implementation}
\kk{We implement our model and perform all algorithm comparisons using PyTorch~\cite{paszke_2017_automatic}. For our model, we use the same ST-GCN backbone and network hyper parameter settings as in~\cite{yan_2018_AAAI}. Network hyper parameters include the number of layers and the initialization of the networks. Other hyperparameters are: number of feature segments $N=5$; number of discriminative features $D=3$; batch size 64; initial learning rate 0.1, which would be divided by 10 in the $60_{th}$ and $90_{th}$ epochs; stochastic gradient descent with Nesterov momentum 0.9; and weight decay 0.0001. The training is ended in the $120_{th}$ epoch.}% We only use  one steam structure to validate our method. It is likely that two stream ensemble models would improve performance the same way as in~\cite{Shi_2019_CVPR2,si_2019_CVPR}.} 

\subsection{Ablation Study}
\subsubsection{\kk{Direction Invariant Features}}
\kk{We test the effectiveness of DIF features with NTU-RGBD 60 and NTU-RGBD 120 datasets using the ST-GCN as the baseline. The first two rows of Table~\ref{tabel:DIF} show that using DIF features can improve the recognition accuracy by a large margin. We thus strongly advocate the usage of DIF features for all skeleton-based action recognition tasks.}

\subsubsection{\kk{Discriminative Feature Learning}}
\kk{The performance gains with discriminative feature learning are shown in the last row of  Table~\ref{tabel:DIF}.}

\begin{table}[htb]
\centering
\caption{Ablation on DIF and DFL.}
\begin{tabular}{|c|l|c|l|c|l|c|l|c|l|}
\hline
\multicolumn{2}{|c|}{Method} & \multicolumn{4}{c|}{NTU-RGBD 60}                                     & \multicolumn{4}{c|}{NTU-RGBD 120}                                     \\ \cline{3-10} 
\multicolumn{2}{|c|}{}           & \multicolumn{2}{c|}{X-View(\%)} & \multicolumn{2}{c|}{X-Subject(\%)} & \multicolumn{2}{c|}{X-Setup(\%)} & \multicolumn{2}{c|}{X-Subject(\%)} \\ \hline
\multicolumn{2}{|c|}{ST-GCN baseline}       & \multicolumn{2}{c|}{88.3}       & \multicolumn{2}{c|}{81.5}          & \multicolumn{2}{c|}{71.3}        & \multicolumn{2}{c|}{72.4}          \\ \hline
\multicolumn{2}{|c|}{ST-GCN + DIF}     & \multicolumn{2}{c|}{92.6}       & \multicolumn{2}{c|}{85.3}          & \multicolumn{2}{c|}{83.4}        & \multicolumn{2}{c|}{81.8}          \\ \hline
\multicolumn{2}{|c|}{ST-GCN + DIF + DFL}                       & \multicolumn{2}{c|}{93.3}       & \multicolumn{2}{c|}{86.7}          & \multicolumn{2}{c|}{85.7}        & \multicolumn{2}{c|}{83.8}          \\ \hline
\end{tabular}
\label{tabel:DIF}
\end{table}

\begin{table}[htb]
\centering
\caption{Ablation study on two-branch fusion}
\begin{tabular}{|c|l|c|l|c|}
\hline
\multicolumn{2}{|c|}{Method}       & \multicolumn{3}{c|}{NTU-RGBD 120}                     \\ \cline{3-5} 
\multicolumn{2}{|c|}{}              & \multicolumn{2}{c|}{X-Subject(\%)} & X-Setup(\%)      \\ \hline
\multicolumn{2}{|c|}{Global Branch Alone} & \multicolumn{2}{c|}{81.8}          & 83.4             \\ \hline
\multicolumn{2}{|c|}{DFL Branch Alone} & \multicolumn{2}{c|}{80.3}          & 81.7             \\ \hline
\multicolumn{2}{|c|}{Two Branches}           & \multicolumn{2}{c|}{83.8}          & 85.7             \\ \hline
\end{tabular}
\label{table:fusion}
\end{table}

\subsubsection{\kk{Fusion of Two Branches}}
\kk{Table~\ref{table:fusion} shows that the classification performance is superior when the global branch and the DFL branch are jointly trained and used in inference together.}

\subsection{\kk{Feature Map Visualization}}
\label{sec:visualization}

\begin{figure}[htb]
\centering
\includegraphics[scale = 0.05]{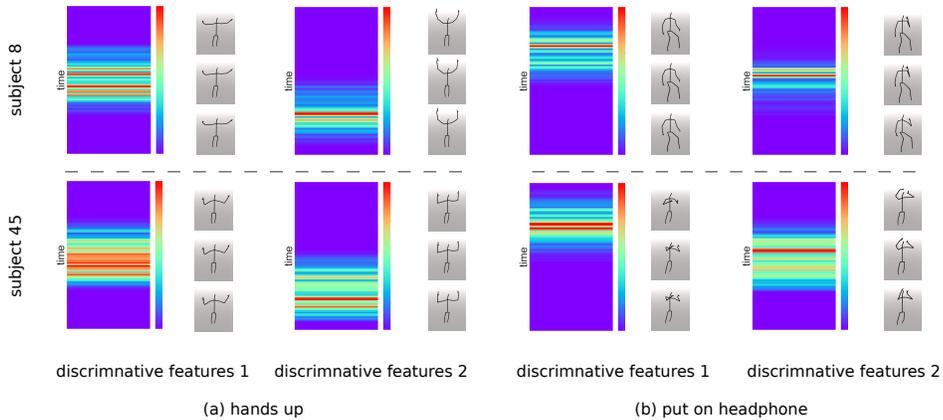}
\caption{Validation of discriminative feature selection. The heat maps show the magnitudes of back propagated gradients for input action clips ``hands up'' (a) and ``put on headphone'' (b). Clip segments with redder colors in the heat maps correspond to the receptive fields of selected discriminative features. We also visualize some sample action frames beside the heat maps to show the motion semantics of these segments. }
\label{align}
\end{figure}

\begin{figure}[htb]
\centering
\includegraphics[scale = 0.07]{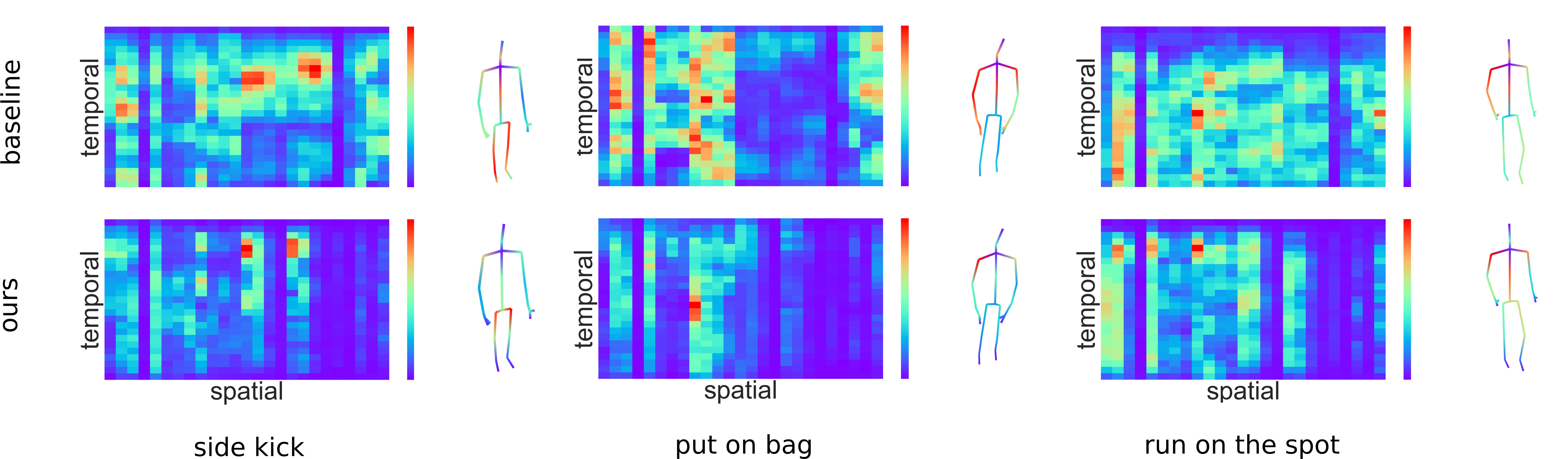}
\caption{Comparisons of feature activations between baseline network and our method. The first line shows the feature map extracted by the baseline backbone network. The second line shows the feature map extracted by our backbone network. The colored character besides the heat map shows corresponding joint activations. Redder colors indicate larger activations.}
\label{feat}
\end{figure}

\kk{To provide more intuition on our discriminative feature selection mechanism, we visualize the receptive fields of selected features by following the visualization methods in~\cite{luo_2016_NIPS}. More specifically, the gradients of the selected discriminative features are back propagated to the input clips. The magnitudes of the gradients are then averaged in the spatial domain. Action segments with large gradients in magnitude correspond to the receptive fields of selected features. For the majority of samples in the datasets, selected features from action clips of the same category are well aligned in time according to their inherent motion semantics, as shown in Figure~\ref{align}.} 

\kk{We also visualize the learned discriminative feature maps and compare them with feature maps learned without the DFL branch. Figure~\ref{feat} shows some examples of our discriminatives feature maps. The activated regions of our feature maps are smaller than those of the baseline method. In addition, they also correspond well with semantically important joints for the input actions. For example, for the ``put on bag'' action, the activations mainly focus around shoulder joints; while the feature map of the baseline method also has large values for other upper body joints. Similarly, for the ``side kick'' action, strong activations of our features focus around the hips. In short, feature maps extracted by our method focus more on semantically important joints, which can help reduce over-fitting and improve the ability to generalize beyond the training data.}

%\begin{table}[tbh]
%\centering
%\caption{Ablation study on inference fusion}
%\begin{tabular}{|c|l|c|l|c|}
%\hline
%\multicolumn{2}{|c|}{Method}       & %\multicolumn{3}{c|}{NTU-RGBD 120}                %     \\ \cline{3-5} 
%\multicolumn{2}{|c|}{}              & %\multicolumn{2}{c|}{X-Subject(\%)} & X-Setup(\%) %     \\ \hline
%\multicolumn{2}{|c|}{Global Branch's outputs} & %\multicolumn{2}{c|}{83.5}          & 85.3         %    \\ \hline
%\multicolumn{2}{|c|}{DFL Branch's outputs} & %\multicolumn{2}{c|}{82.1}          & 83.7        %     \\ \hline
%\multicolumn{2}{|c|}{Two Branches}           & %\multicolumn{2}{c|}{83.8}          & 85.7        %     \\ \hline
%\end{tabular}
%\label{table:fusionInference}
%\end{table}

\subsection{Generalization and Robustness}

\begin{figure}[htb]
  \begin{minipage}[c]{0.58\textwidth}
  \centering
    \includegraphics[scale = 0.028]{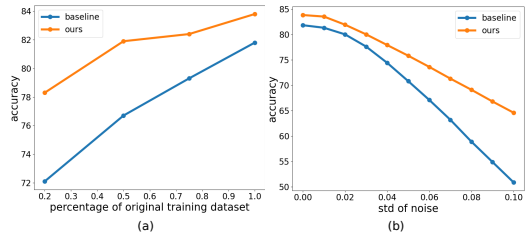}
  \end{minipage}\hfill
  \begin{minipage}[c]{0.40\textwidth}
  \centering
    \caption{
       We test the robustness of our model on the cross-subject benchmark of NTU-RGBD 120. (a) Classification accuracy with reduced training data. (b) Classification accuracy with noisy inputs.
    } \label{noise}
  \end{minipage}
\end{figure}

\kk{We validate the generalization ability of our classification network by using less training data. In the original NTU-RGBD 120 cross-subject benchmark, 53 subjects are used for training and the other 53 for testing. We keep the test dataset unchanged and gradually shrink the size of the training data. We plot the recognition accuracy with respect to the size of the training data in Figure~\ref{noise}(a). The performance degrades more gracefully using our method compared with the baseline method. Good generalization is crucial in applications where training data is hard to collect.}

%\begin{figure}[htb]
%\centering
%\includegraphics[scale = 0.03]{images/noise&few.pdf}
%\caption{We test the robustness of our model on the cross-subject benchmark of NTU-RGBD 120. (a) Classification accuracy with reduced training data. (b) Classification accuracy with noisy inputs.}
%\label{noise}
%\end{figure}

\kk{We validate the robustness of our model by adding noise to the input clips. Most captured 3D human skeleton datasets are clean and of good quality, while in real-world applications, the input data may be much noisier. We randomly add Gaussian noise to all joints with the standard deviation ranging from $0$ to $0.1$ meters. Figure~\ref{noise}(b) shows that our network is more robust to noise at all levels, compared with the baseline method.}

\begin{table}[htb]
\centering
\caption{Comparison with ESA method}
\begin{tabular}{|c|l|c|l|c|}
\hline
\multicolumn{2}{|c|}{Method}                  & \multicolumn{3}{c|}{NTU-RGBD 120}                     \\ \cline{3-5} 
\multicolumn{2}{|c|}{}                    & \multicolumn{2}{c|}{X-Subject(\%)} & X-Setup(\%)      \\ \hline
\multicolumn{2}{|c|}{ST-GCN}                                 & \multicolumn{2}{c|}{81.8}          & 83.4             \\ \hline
\multicolumn{2}{|c|}{ST-GCN + ESA} & \multicolumn{2}{c|}{82.1}          & 83.6             \\ \hline
\multicolumn{2}{|c|}{Ours}                                     & \multicolumn{2}{c|}{83.8}          & 85.7             \\ \hline
\end{tabular}
\label{table: spatial attention}
\end{table}

\subsection{\kk{Comparison with Explicit Spatial Attention Method}}
\kk{Since our method can help the GCN backbone to focus on spatially more important joints, we also compare it with the Explicit Spatial Attention (ESA) method that can roughly achieve the same effect~\cite{hu_2018_NIPS,vaswani_2017_NIPS,wang_2017_CVPR,xu_2015_ICML,shi_2019_arxiv}. More specifically, for each spatial-temporal graph convolution block, we compute an attention mask $\mathbf{M}_{s}$ as follows:
\begin{equation}
    \mathbf{M}_{s} = \sigma(\mathbf{C}_{s}(\mathbf{P}_t(\mathbf{f}_{out})))
\end{equation}
where $\mathbf{f}_{out}$ is the feature map after each block. $\mathbf{P}_t$ is a pooling layer that averages $\mathbf{f}_{out}$ in the temporal domain. $\mathbf{C}_s$ is a 1D convolution in the spatial domain. $\sigma$ is the Sigmoid function. The output feature is then computed as:
\begin{equation}
    \mathbf{f}_{out} = \mathbf{f}_{out} + \mathbf{M}_{s} * \mathbf{f}_{out}
\end{equation}}

\kk{Table~\ref{table: spatial attention} shows that with the above ESA method, the recognition accuracy can be improved to some degree, but still not comparable with our method. The ESA method could be interpreted as an addition mechanism, while our method can be seen as a regularization or substraction mechanism. This is because our model does not change the backbone architecture, and the DFL branch helps regularize the GCN backbone to learn more robust features in the spatial domain.}

\begin{table}[htb]
\parbox{.39\linewidth}
{\centering
\caption{Comparison on SYSU}
\begin{tabular}{cc}
\hline
Method                          & Accuracy \\ \hline
D-Skeleton \cite{hu_2015_CVPR}                       & 75.5     \\
ST-LSTM \cite{liu_2017_TPAMI}                         & 76.5     \\
DPRL\cite{tang_2018_ICCV}                            & 76.9     \\
SR-TSL\cite{si_2018_ECCV}                          & 80.7     \\
BGS-LSTM\cite{zhao_2019_ICCV} & \textbf{82.0}     \\ \hline
Ours                            & 81.7     \\ \hline
\end{tabular}
\label{table:SYSU}}
\hfill
\parbox{.59\linewidth}
{\centering
\caption{Comparison on Kinectics-Skeleton}
\begin{tabular}{ccllcll}
\hline
Methods     & \multicolumn{3}{c}{Top-1(\%)} & \multicolumn{3}{c}{Top-5(\%)} \\ \hline
Feature Enc\cite{Fernando_2015_CVPR} & \multicolumn{3}{c}{14.9}      & \multicolumn{3}{c}{25.8}      \\
Deep LSTM\cite{Shahroudy_2016_CVPR}   & \multicolumn{3}{c}{16.4}      & \multicolumn{3}{c}{35.3}      \\
TCN\cite{kim_2017_CVPRW}         & \multicolumn{3}{c}{20.3}      & \multicolumn{3}{c}{40.0}      \\
ST-GCN\cite{yan_2018_AAAI}      & \multicolumn{3}{c}{30.7}      & \multicolumn{3}{c}{52.8}      \\
AS-GCN\cite{Li_2019_CVPR}      & \multicolumn{3}{c}{34.8}      & \multicolumn{3}{c}{56.5}      \\
Js-AGCN\cite{Shi_2019_CVPR2}     & \multicolumn{3}{c}{\textbf{35.1}}      & \multicolumn{3}{c}{\textbf{57.1}}      \\ \hline
%DGNN\cite{Shi_2019_CVPR}        & \multicolumn{3}{c}{\textbf{36.9}}      & \multicolumn{3}{c}{\textbf{59.6}}      \\ \hline
Ours        & \multicolumn{3}{c}{32.1}      & \multicolumn{3}{c}{54.2}      \\ \hline
\end{tabular}
\label{table:kinetics}
}
\end{table}

%%%%%%%%%%%%%%%%%%%%%%%%%%%%%%%%%%%%%%%%%%%%%%%%
\begin{table}[htb]
\centering
\caption{Comparison on NTU-RGBD 60. (`-' indicates no data available.)}
\begin{tabular}{@{}lllllll@{}}
\hline
Methods                 & \multicolumn{3}{c}{X-Sub(\%)} & \multicolumn{3}{c}{X-View(\%)} \\ \hline
%Lie Group\cite{vemulapall_2014_CVPR}               & \multicolumn{3}{c}{50.1}      & \multicolumn{3}{c}{82.8}       \\ \hline
%HBRNN\cite{du_2015_CVPR}                   & \multicolumn{3}{c}{59.1}      & \multicolumn{3}{c}{64.0}       \\
%Deep LSTM\cite{Shahroudy_2016_CVPR}         & \multicolumn{3}{c}{60.7}      & \multicolumn{3}{c}{67.3}       \\
%ST-LSTM\cite{liu_2016_ECCV}                 & \multicolumn{3}{c}{69.2}      & \multicolumn{3}{c}{77.7}       \\
STA-LSTM\cite{song_2017_AAAI}                & \multicolumn{3}{c}{73.4}      & \multicolumn{3}{c}{81.2}       \\
VA-LSTM\cite{zhang_2017_CVPR}                 & \multicolumn{3}{c}{79.2}      & \multicolumn{3}{c}{87.7}       \\
ARRN-LSTM\cite{li_2018_arxiv_skeleton}               & \multicolumn{3}{c}{80.7}      & \multicolumn{3}{c}{88.8}       \\
Ind-RNN\cite{li_2018_CVPR}               & \multicolumn{3}{c}{81.8}      & \multicolumn{3}{c}{88.0}       \\
AGC-LSTM(joint  stream)\cite{si_2019_CVPR} & \multicolumn{3}{c}{87.5}      & \multicolumn{3}{c}{93.5}       \\ \hline
TCN\cite{kim_2017_CVPRW}                     & \multicolumn{3}{c}{74.3}      & \multicolumn{3}{c}{83.1}       \\
Clips+CNN+MTLN\cite{ke_2017_CVPR}          & \multicolumn{3}{c}{79.6}      & \multicolumn{3}{c}{84.8}       \\
Synthesized CNN\cite{liu_2017_PR}         & \multicolumn{3}{c}{80.0}      & \multicolumn{3}{c}{87.2}       \\
3scale ResNet152\cite{li_2017_ICMEW}       & \multicolumn{3}{c}{85.0}      & \multicolumn{3}{c}{92.3}       \\ \hline
ST-GCN\cite{yan_2018_AAAI}                  & \multicolumn{3}{c}{81.5}      & \multicolumn{3}{c}{88.3}       \\
Js-AGCN\cite{Shi_2019_CVPR2}                 & \multicolumn{3}{c}{-}         & \multicolumn{3}{c}{93.7}       \\
%DGNN\cite{Shi_2019_CVPR}                    & \multicolumn{3}{c}{\textbf{89.9}}      & \multicolumn{3}{c}{\textbf{96.1}}       \\
AS-GCN\cite{Li_2019_CVPR}                  & \multicolumn{3}{c}{\textbf{86.8}}      & \multicolumn{3}{c}{\textbf{94.2}}       \\\hline
%ST-GCN(DIF)             & \multicolumn{3}{c}{85.3}      & \multicolumn{3}{c}{92.6}       \\ \hline
Ours                    & \multicolumn{3}{c}{86.7}      & \multicolumn{3}{c}{93.3}       \\ \hline
\end{tabular}
\label{table:NTU_60}
\end{table}

\subsection{\kk{Comparison with State-of-the-Art Methods}}
\kk{We conduct comparative studies on four datasets described in Section~\ref{sec:datasets}. For SYSU and NTU-RGBD 60 datasets, we achieve comparable results as shown in Table~\ref{table:SYSU} and \ref{table:NTU_60}. The accuracy for other methods are taken from the reference papers directly. For NTU-RGBD 120, our model achieve better results by a large margin as shown in Table~\ref{table:NTU_120}. The accuracy for other methods are either taken from the reference papers directly or obtained by running author-released code (ST-GCN,Js-AGCN,AS-GCN). These tests manifest that the performance of our method degrades more gracefully than other methods when moving from small datasets to larger and more challenging datasets. We also note that our model is based on the original ST-GCN architecture. Compared with other improvements of ST-GCN such as AS-GCN and Js-AGCN, our model has much smaller size as shown in Table~ \ref{table:NTU_120}.}%{Table \ref{table:fusion} shows that classification accuracy using only global branch is higher than Js-AGCN and AS-GCN, which means our method can be used to train models with fewer parameters while better performance. }

%%%%%%%%%%%%%%%%%%%%%%%%%%%%%%%%%%%%%%%%%%%%%%%%%%%%%%%%%%%%%%%%%%%%%%%%%%%%%%%%%%%%%%%%
\begin{table}[tbh]
\centering
\caption{Comparison on NTU-RGBD 120.(`-' indicates no data available for the method. `*' indicates runtime crash from author-released code.)}
\begin{tabular}{ccllcllcllc}
\hline
Methods                                                           & \multicolumn{3}{c}{X-Sub(\%)}     & \multicolumn{3}{c}{X-View(\%)}    & \multicolumn{3}{c}{X-Setup(\%)}   & \multicolumn{1}{l}{model size} \\ \hline
SkeleMotion\cite{Caetano_2019_AVSS}            & \multicolumn{3}{c}{67.7}          & \multicolumn{3}{c}{-}             & \multicolumn{3}{c}{66.9}          & -                              \\
Body Pose Evolution Map\cite{liu_2018_CVPR}    & \multicolumn{3}{c}{64.6}          & \multicolumn{3}{c}{-}             & \multicolumn{3}{c}{66.9}          & -                              \\
Multi-Task CNN with RotClips\cite{ke_2018_TIP} & \multicolumn{3}{c}{62.2}          & \multicolumn{3}{c}{-}             & \multicolumn{3}{c}{61.8}          & -                              \\
Two-Stream Attention LSTM\cite{liu_2017_TPAMI} & \multicolumn{3}{c}{61.2}          & \multicolumn{3}{c}{-}             & \multicolumn{3}{c}{63.3}          & -                              \\
Synthesized CNN\cite{liu_2017_PR}              & \multicolumn{3}{c}{60.3}          & \multicolumn{3}{c}{-}             & \multicolumn{3}{c}{63.2}          & -                              \\
Clips+CNN+MTLN\cite{ke_2017_CVPR}              & \multicolumn{3}{c}{58.4}          & \multicolumn{3}{c}{-}             & \multicolumn{3}{c}{57.9}          & -                              \\ \hline
ST-GCN\cite{yan_2018_AAAI}                     & \multicolumn{3}{c}{72.4}          & \multicolumn{3}{c}{81.2}          & \multicolumn{3}{c}{71.3}          & \textbf{5.8MB}                          \\
AS-GCN\cite{Li_2019_CVPR}                      & \multicolumn{3}{c}{77.7}          & \multicolumn{3}{c}{*}             & \multicolumn{3}{c}{79.0}          & 28.4MB                         \\
Js-AGCN\cite{Shi_2019_CVPR2}                   & \multicolumn{3}{c}{82.8}          & \multicolumn{3}{c}{90.9}          & \multicolumn{3}{c}{84.4}          & 14.9MB                         \\ \hline
Ours                                                              & \multicolumn{3}{c}{\textbf{83.8}} & \multicolumn{3}{c}{\textbf{91.3}} & \multicolumn{3}{c}{\textbf{85.7}} & 6.4MB                 \\ \hline
\end{tabular}
\label{table:NTU_120}
\end{table}

\kk{We also test the performance on Kinectics-Skeleton. As Table~\ref{table:kinetics} shows, all methods perform poorly.%, with DGNN~\cite{Shi_2019_CVPR} scores the highest. We note that DGNN is a two-stream method that utilizes both joint positions and bone locations at the feature level, while all other methods in the table only use joint positions. 
The challenges with Kinectics-Skeleton include that extracted 2D joint positions are of very low quality, and there are action classes which are indistinguishable from just skeletal poses, such as ``eating hot dog'' and ``eating burger''. In the future, we plan to develop an integrated model that utilizes features from both skeletal poses and images to further boost the performance.} %It might be caused by the data modality of the dataset. Skeletons of this dataset only contains 2D positions of joints with large noises since they are extracted from images by OpenPose. In our experiments we found that GCN based models over-fit training dataset where skeletons are captured by 3D sensors, while they under-fit in Skeleton-Kinetics dataset. Our method is designed for avoid over-fitting, so it is reasonable that our method can't outperform other methods in this dataset. \comm{Besides the data quality issue, there are many action classes which are hard to recognize by using only skeleton information like "eating hot dog" and "eating burger". In future work, we plan to fuse features from skeleton and images to extract more essential and robust features to improve action recognition accuracy.}}

\section{Conclusions}
\kk{We have proposed a neural network architecture which facilitates the GCN backbone to learn more discriminative spatial and temporal features for skeleton-based action recognition tasks. This model is simple to implement and effective on all skeleton datasets of reasonable quality. Our ablation studies and experimental results on four datasets testify the effectiveness of our method.}

\kk{In addition, direction invariant features, which can be simply extracted in a preprocessing stage, can greatly improve the performance of state-of-the-art models. No previous work has used DIF features in a principled way, to the best of our knowledge.}

\kk{Our proposed method still cannot generate acceptable results for low-quality skeletons extracted from Kinetics-Skeleton, similar to all other state-of-the-art methods. For future work, we plan to integrate our skeleton-based method with video-based methods to tackle action recognition on such challenging datasets.}

\bibliographystyle{splncs04}
\bibliography{action}

\begin{thebibliography}{10}
\providecommand{\url}[1]{\texttt{#1}}
\providecommand{\urlprefix}{URL }
\providecommand{\doi}[1]{https://doi.org/#1}

\bibitem{Caetano_2019_AVSS}
Caetano, C., Sena, J., Bremond, F., Dos~Santos, J.A., Schwartz, W.R.:
  Skelemotion: A new representation of skeleton joint sequences based on motion
  information for {3D} action recognition. AVSS  (2019)

\bibitem{Cao_2017_CVPR}
Cao, Z., Simon, T., Wei, S.E., Sheikh, Y.: Realtime multi-person {2D} pose
  estimation using part affinity fields. CVPR  (2017)

\bibitem{Fernando_2015_CVPR}
Fernando, B., Gavves, E., Oramas, J.M., Ghodrati, A., Tuytelaars, T.: Modeling
  video evolution for action recognition. In: CVPR (2015)

\bibitem{hu_2015_CVPR}
Hu, J.F., Zheng, W.S., Lai, J., Zhang, J.: Jointly learning heterogeneous
  features for {RGB-D} activity recognition. In: CVPR (2015)

\bibitem{hu_2018_NIPS}
Hu, J., Shen, L., Albanie, S., Sun, G., Vedaldi, A.: Gather-excite: Exploiting
  feature context in convolutional neural networks. In: NeuralIPS (2018)

\bibitem{kay_2017_kinetics}
Kay, W., Carreira, J., Simonyan, K., Zhang, B., Hillier, C., Vijayanarasimhan,
  S., Viola, F., Green, T., Back, T., Natsev, P., et~al.: The kinetics human
  action video dataset. arXiv preprint arXiv:1705.06950  (2017)

\bibitem{ke_2017_CVPR}
Ke, Q., Bennamoun, M., An, S., Sohel, F., Boussaid, F.: A new representation of
  skeleton sequences for {3D} action recognition. In: CVPR (2017)

\bibitem{ke_2018_TIP}
Ke, Q., Bennamoun, M., An, S., Sohel, F., Boussaid, F.: Learning clip
  representations for skeleton-based {3D} action recognition. TIP  (2018)

\bibitem{kim_2017_CVPRW}
Kim, T.S., Reiter, A.: Interpretable {3D} human action analysis with temporal
  convolutional networks. In: CVPRW (2017)

\bibitem{li_2017_ICMEW}
Li, B., Dai, Y., Cheng, X., Chen, H., Lin, Y., He, M.: Skeleton based action
  recognition using translation-scale invariant image mapping and multi-scale
  deep {CNN}. In: ICMEW (2017)

\bibitem{li_2017_ICMEW_2}
Li, C., Zhong, Q., Xie, D., Pu, S.: Skeleton-based action recognition with
  convolutional neural networks. In: ICMEW (2017)

\bibitem{li_2018_arxiv_co}
Li, C., Zhong, Q., Xie, D., Pu, S.: Co-occurrence feature learning from
  skeleton data for action recognition and detection with hierarchical
  aggregation. arXiv preprint arXiv:1804.06055  (2018)

\bibitem{li_2018_arxiv_skeleton}
Li, L., Zheng, W., Zhang, Z., Huang, Y., Wang, L.: Skeleton-based relational
  modeling for action recognition. arXiv preprint arXiv:1805.02556  (2018)

\bibitem{Li_2019_CVPR}
Li, M., Chen, S., Chen, X., Zhang, Y., Wang, Y., Tian, Q.: Actional-structural
  graph convolutional networks for skeleton-based action recognition. In: CVPR
  (2019)

\bibitem{li_2019_arxiv_sym}
Li, M., Chen, S., Chen, X., Zhang, Y., Wang, Y., Tian, Q.: Symbiotic graph
  neural networks for {3D} skeleton-based human action recognition and motion
  prediction. arXiv preprint arXiv:1910.02212  (2019)

\bibitem{li_2018_CVPR}
Li, S., Li, W., Cook, C., Zhu, C., Gao, Y.: Independently recurrent neural
  network ({INDRNN}): Building a longer and deeper {RNN}. In: CVPR (2018)

\bibitem{liu_2017_arxiv}
Liu, H., Tu, J., Liu, M.: Two-stream 3{D} convolutional neural network for
  skeleton-based action recognition. arXiv preprint arXiv:1705.08106  (2017)

\bibitem{liu_2019_TPAMI}
Liu, J., Shahroudy, A., Perez, M.L., Wang, G., Duan, L.Y., Chichung, A.K.: {NTU
  RGB+D 120}: A large-scale benchmark for {3D} human activity understanding.
  TPAMI  (2019)

\bibitem{liu_2017_TPAMI}
Liu, J., Shahroudy, A., Xu, D., Kot, A.C., Wang, G.: Skeleton-based action
  recognition using spatio-temporal {LSTM} network with trust gates. TPAMI
  (2017)

\bibitem{liu_2016_ECCV}
Liu, J., Shahroudy, A., Xu, D., Wang, G.: Spatio-temporal {LSTM} with trust
  gates for {3D} human action recognition. In: ECCV (2016)

\bibitem{liu_2017_PR}
Liu, M., Liu, H., Chen, C.: Enhanced skeleton visualization for view invariant
  human action recognition. PR  (2017)

\bibitem{liu_2018_CVPR}
Liu, M., Yuan, J.: Recognizing human actions as the evolution of pose
  estimation maps. In: CVPR (2018)

\bibitem{luo_2016_NIPS}
Luo, W., Li, Y., Urtasun, R., Zemel, R.: Understanding the effective receptive
  field in deep convolutional neural networks. In: NeuralIPS (2016)

\bibitem{Ma_2019_PG}
Ma, L.K., Yang, Z., Guo, B., Yin, K.: Towards robust direction invariance in
  character animation. Computer Graphics Forum  (2019)

\bibitem{martinez_2019_ICCV}
Martinez, B., Modolo, D., Xiong, Y., Tighe, J.: Action recognition with
  spatial-temporal discriminative filter banks. In: ICCV (2019)

\bibitem{paszke_2017_automatic}
Paszke, A., Gross, S., Chintala, S., Chanan, G., Yang, E., DeVito, Z., Lin, Z.,
  Desmaison, A., Antiga, L., Lerer, A.: Automatic differentiation in pytorch
  (2017)

\bibitem{peng_2019_AAAI}
Peng, W., Hong, X., Chen, H., Zhao, G.: Learning graph convolutional network
  for skeleton-based human action recognition by neural searching. arXiv
  preprint arXiv:1911.04131  (2019)

\bibitem{Shahroudy_2016_CVPR}
Shahroudy, A., Liu, J., Ng, T.T., Wang, G.: {NTU RGB+D}: A large scale dataset
  for {3D} human activity analysis. In: CVPR (2016)

\bibitem{Shi_2019_CVPR}
Shi, L., Zhang, Y., Cheng, J., Lu, H.: Skeleton-based action recognition with
  directed graph neural networks. In: CVPR (2019)

\bibitem{shi_2019_arxiv}
Shi, L., Zhang, Y., Cheng, J., LU, H.: Skeleton-based action recognition with
  multi-stream adaptive graph convolutional networks. arXiv preprint
  arXiv:1912.06971  (2019)

\bibitem{Shi_2019_CVPR2}
Shi, L., Zhang, Y., Cheng, J., Lu, H.: Two-stream adaptive graph convolutional
  networks for skeleton-based action recognition. In: CVPR (2019)

\bibitem{si_2019_CVPR}
Si, C., Chen, W., Wang, W., Wang, L., Tan, T.: An attention enhanced graph
  convolutional {LSTM} network for skeleton-based action recognition. In: CVPR
  (2019)

\bibitem{si_2018_ECCV}
Si, C., Jing, Y., Wang, W., Wang, L., Tan, T.: Skeleton-based action
  recognition with spatial reasoning and temporal stack learning. In: ECCV
  (2018)

\bibitem{song_2017_AAAI}
Song, S., Lan, C., Xing, J., Zeng, W., Liu, J.: An end-to-end spatio-temporal
  attention model for human action recognition from skeleton data. In: AAAI
  (2017)

\bibitem{sun_2018_ECCV}
Sun, Y., Zheng, L., Yang, Y., Tian, Q., Wang, S.: Beyond part models: Person
  retrieval with refined part pooling (and a strong convolutional baseline).
  In: ECCV (2018)

\bibitem{tang_2018_ICCV}
Tang, Y., Tian, Y., Lu, J., Li, P., Zhou, J.: Deep progressive reinforcement
  learning for skeleton-based action recognition. In: CVPR (2018)

\bibitem{vaswani_2017_NIPS}
Vaswani, A., Shazeer, N., Parmar, N., Uszkoreit, J., Jones, L., Gomez, A.N.,
  Kaiser, {\L}., Polosukhin, I.: Attention is all you need. In: NeuralIPS
  (2017)

\bibitem{vemulapall_2014_CVPR}
Vemulapalli, R., Arrate, F., Chellappa, R.: Human action recognition by
  representing {3D} skeletons as points in a lie group. In: CVPR (2014)

\bibitem{wang_2017_CVPR}
Wang, F., Jiang, M., Qian, C., Yang, S., Li, C., Zhang, H., Wang, X., Tang, X.:
  Residual attention network for image classification. In: CVPR (2017)

\bibitem{wang_2016_ECCV}
Wang, L., Xiong, Y., Wang, Z., Qiao, Y., Lin, D., Tang, X., Van~Gool, L.:
  Temporal segment networks: Towards good practices for deep action
  recognition. In: ECCV (2016)

\bibitem{xu_2015_ICML}
Xu, K., Ba, J., Kiros, R., Cho, K., Courville, A., Salakhudinov, R., Zemel, R.,
  Bengio, Y.: Show, attend and tell: Neural image caption generation with
  visual attention. In: ICML (2015)

\bibitem{yan_2018_AAAI}
Yan, S., Xiong, Y., Lin, D.: Spatial temporal graph convolutional networks for
  skeleton-based action recognition. In: AAAI (2018)

\bibitem{zhang_2017_CVPR}
Zhang, P., Lan, C., Xing, J., Zeng, W., Xue, J., Zheng, N.: View adaptive
  recurrent neural networks for high performance human action recognition from
  skeleton data. In: ICCV (2017)

\bibitem{zhang_2017_arxiv}
Zhang, X., Luo, H., Fan, X., Xiang, W., Sun, Y., Xiao, Q., Jiang, W., Zhang,
  C., Sun, J.: Alignedreid: Surpassing human-level performance in person
  re-identification. arXiv preprint arXiv:1711.08184  (2017)

\bibitem{zhao_2019_ICCV}
Zhao, R., Wang, K., Su, H., Ji, Q.: Bayesian graph convolution {LSTM} for
  skeleton based action recognition. In: ICCV (2019)

\end{thebibliography}
\end{document}